# Early screening of potential breakthrough technologies with enhanced interpretability: A patent-specific hierarchical attention network model

*Jaewoong Choi[a], Janghyeok Yoon[b], Changyong Lee[c,*]*

[a] Computational Science Research Center, Korea Institute of Science and Technology, 5 Hwarang-ro, Seongbuk-gu, Seoul 02792, Republic of Korea

[b] Department of Industrial Engineering, Konkuk University, 120 Neungdong-ro, Gwangjin-gu, Seoul 05029, Republic of Korea

[c] Department of Public Administration, Korea University, 145 Anam-ro, Seongbuk-gu, Seoul 02841, Republic of Korea

[*] Corresponding author: Changyong Lee (changyonglee@korea.ac.kr)

**Abstract**

Despite the usefulness of machine learning approaches for the early screening of potential breakthrough technologies, their practicality is often hindered by opaque models. To address this, we propose an interpretable machine learning approach to predicting future citation counts from patent texts using a patent-specific hierarchical attention network (PatentHAN) model. Central to this approach are (1) a patent-specific pre-trained language model, capturing the meanings of technical words in patent claims, (2) a hierarchical network structure, enabling detailed analysis at the claim level, and (3) a claim-wise self-attention mechanism, revealing pivotal claims during the screening process. A case study of 35,376 pharmaceutical patents demonstrates the effectiveness of our approach in early screening of potential breakthrough technologies while ensuring interpretability. Furthermore, we conduct additional analyses using different language models and claim types to examine the robustness of the approach. It is expected that the proposed approach will enhance expert-machine collaboration in identifying breakthrough technologies, providing new insight derived from text mining into technological value.

# 1. Introduction

Breakthrough technologies have received significant attention from both industry and academia as key enablers of sustainable growth and competitive edge. Prior studies have presented a range of approaches, from qualitative methods such as Delphi and SWOT to quantitative models (e.g., stochastic models and machine learning models), utilizing diverse data sources such as experts' opinions, patents, scientific publications, and technology news and blogs (Lee, 2021). Among others, machine learning (ML) models integrated with patent databases are acknowledged for their effectiveness in identifying potential breakthrough technologies. This is attributed to the comprehensive coverage and structured information of patent databases, coupled with the predictive capabilities of ML models to analyze large datasets and uncover novel insights (Kim et al., 2019; Lee et al., 2018). However, the practicality of this approach is often hindered by the opacity of ML models (Kim et al., 2022), particularly when evaluating early-stage technological ideas (Hong et al., 2022).

Previous patent-based ML approaches to identifying potential breakthrough technologies can be classified into two categories according to the input data used: bibliometric and textual information. The former employs ML models to associate ex-ante patent quality indicators as inputs and patent forward citations as outputs, to predict future citation counts of relevant patents as a proxy of technological value. Although this approach has further been extended to interpretable models and proven useful in various contexts (Kim et al., 2022), it can only be performed at later stages of technology development because it requires patent information such as patent family and classification codes available at the time or after a patent is granted. The latter leverages ML models to examine the relationships between textual information of patents as inputs and patent forward citations as outputs. This approach has proven effective in evaluating early-stage technological ideas, where ex-ante patent quality indicators are unavailable but technical descriptions of ideas exist (Hong et al., 2022; Woo et al., 2019). However, this approach relies on black-box models, underscoring the need to develop interpretable models that enable experts to grasp the underlying mechanisms of prediction processes.

Developing interpretable ML models with textual information of patents presents several challenges. Text data is inherently high-dimensional and sparse, typically represented by a large number of features (e.g., words and tokens). Although text data can be transformed into dense embeddings using pre-trained language models (PLMs) such as word2vec and BERT, which capture rich linguistic patterns and meanings, interpreting the specific dimensions or components within dense embeddings can be non-trivial, as they often represent latent features derived from large amounts of textual data. Moreover, their high dimensionality and complex relationships make it challenging to interpret how individual features contribute to machine predictions. Furthermore, the technical content of patents can be more efficiently interpreted on a claim-wise basis. Beyond the level of individual words and sentences, it is necessary to reflect such specific claim structures in the model architecture.

Considering these issues, we propose an interpretable ML approach to predicting future citation counts from patent texts, reporting a patent-specific hierarchical attention network (PatentHAN) model. At the heart of this model are: (1) PatentBERT, a patent-specific PLM, is employed in a text encoding layer, which transforms the patent claim texts into dense embedding vectors considering complex meanings of technical contents; (2) a hierarchical network structure, where individual claim vectors are structured to generate a patent representation vector, is designed to conduct claim-level analysis; (3) a claim-wise self-attention mechanism, which assigns weights to each claim vector and allows the model to focus more on semantically significant parts, is adopted to identify pivotal claims during the screening process. This self-attention mechanism naturally captures the interactions between inputs (i.e., claim vectors), which enables the model to weigh the significance of each claim appropriately during training during the model's training process. As a result, the outcomes, such as claim-wise attention scores, effectively evaluate the relative importance of multiple claims. Once trained, PatentHAN discriminates whether the given patent is a potential breakthrough technology from its textual information. At the same time, the claim-wise attention mechanism reveals the pivotal claims with high attention weights.

We applied the proposed approach to 35,376 patents in the pharmaceutical technology field. Our case study confirmed that the proposed approach is effective in screening potential breakthrough technologies and the claim-wise attention mechanism enables us to interpret the prediction by revealing the most pivotal claim. Also, we verified robust performance with around 89% accuracy in various contexts by testing multiple PLMs and found that the PLMs trained on scientific literature performed slightly better than PatentBERT as domain knowledge is required to understand the contents of pharmaceutical technologies. We found that the attention scores of dependent claims were statistically significantly lower than those of independent claims. This finding aligns with previous research, which indicates that independent claims hold greater importance compared to dependent claims (Marco et al., 2019; Reitzig, 2004). While the outcomes were not entirely conclusive, they indirectly validated our model and demonstrated a practical pathway for implementing the proposed approach. It is expected that the proposed approach will make practical contributions by effectively supporting expert-machine collaboration in identifying breakthrough technologies, while the quantitative outcomes have academic implications by suggesting new text-mined insights into technological value.

The remainder of this paper is organized as follows. Section 2 presents the summary of previous works and Section 3 explains the proposed approach, which is illustrated with a case study in Section 4. Section 5 discusses the practical implications of the proposed approach and, finally, Section 6 offers our conclusions.

## 2. Background

Breakthrough technologies have been analyzed with patent citation information, as it signifies the impact of patents on follow-up technologies. In this regard, many researchers have suggested patent-based approaches to estimate the potential of breakthrough technologies by predicting the number of forward citations. Initially, there have been scientific methods such as curve fitting methods and stochastic processes to provide forward-looking insights on potential breakthrough technologies, by estimating future citation count of relevant patents. Shin et al. (2013) adopted curve-fitting techniques to estimate both the expected number of patent citations and its variance, as a proxy for assessing the future returns and risks associated with technology of interest. Some researchers introduced a stochastic patent citation analysis method such as Pareto, Negative binomial distribution model, or Hawkes process to estimate the future impacts of a period of interest (Jang et al., 2017; Lee et al., 2012). Lee et al. (2016) and Lee et al. (2017) suggested a stochastic technology life cycle analysis, utilizing hidden Markov models and time-series patent indicators, to study the technological progression over its life cycle. Although previous studies have been valuable in providing quantitative outcomes and scientific methods regarding breakthrough technologies, their applicability is limited when relevant technologies are at the early stages of technology development without sufficient historical citation data (Lee et al., 2018).

As a remedy, several researchers have developed ML models to identify potential breakthrough technologies at the early stages of development by predicting the future values of individual patents. For instance, Lee et al. (2018) utilized artificial neural networks to capture the nonlinear relationships between technological values and ex-ante patent indicators that can be determined immediately after relevant patents are issued. Specifically, they employed 18 patent indicators representing technological characteristics − such as novelty, scientific intensity, growth speed, technological scope, and patent development efforts and capabilities − as input variables. However, these approaches, which rely on patent information as their inputs, are impractical for screening potential breakthrough technologies in the early stages of technology development, especially when technologies are at the proof-of-concept or prototype stage. Such methods are typically applied after relevant technologies have been patented, as they rely on information available upon patent registration, such as patent families or scientific references.

In this context, recent studies have aimed to accelerate the screening of potential breakthrough technologies by using patent text as inputs (Hong et al., 2022). These approaches have developed ML models to associate patent text − representing the technical contents of relevant technologies − and the number of forward citations, indicative of the potential technological values. For example, some researchers represented patent texts as keyword vectors and used them as an input of the ML model to predict future citation counts (Woo et al., 2019). Recently, other researchers have applied word2vec and convolutional neural network models to patent abstract text to capture more distinct technical meanings and effectively model the relationship with patent citation counts (Hong et al., 2022). These approaches leverage patent texts as input, allowing for the screening of breakthrough technologies before patenting. By assessing proof-of-concept or prototype-stage technologies, they significantly accelerate the screening process, enhancing applicability in rapidly evolving technology domains.

Despite the usefulness of ML approaches, their practicality is limited due to their black-box nature. Particularly, when machine predictions and experts' judgments are different, the opacity may decrease the effectiveness and increase confusion as there is no way to know the inner mechanism. The interpretability of ML models using ex-ante patent indicators as inputs has indeed been investigated with Shapley additive explanations (SHAP) values in several studies (Kim et al., 2022; Yao and Ni, 2023). However, there has been a notable absence of such attempts to interpret ML models dealing with text inputs. This challenge arises because text data is inherently high-dimensional and sparse. To address this, text data is transformed into dense embedding vectors using PLMs, capturing its semantic meanings. However, each dimension of these text embeddings is a latent vector, making it difficult to directly associate prediction results with the meaning of the text. Existing interpretation techniques, such as SHAP, which determine the significance of individual features, are not readily applicable to textual data. Therefore, structuring text into word-level or sentence-level unit vectors and analyzing their influence can be more effective in interpreting models than using document-level latent vectors. In the case of patents, claim-level analysis can be more effective than word or sentence-level in understanding the relationships between technical elements and technological value. These challenges serve as our primary motivation and are thoroughly addressed in this study. Table 1 summarizes the difference between previous research and the proposed approach.

**Table 1.** Comparison of previous research and the proposed approach

| Approach | Input | Method | Time available | Interpretability |
|---|---|---|---|---|
| Probabilistic approach | Patent citation patterns | Curve fitting techniques; Stochastic | Later stages of technology development | Not provided |

| | | models | | |
|---|---|---|---|---|
| ML approach | Ex-ante patent indicators | ML models | Early stages of technology development – patent information is available | The importance of each patent indicator can be obtained from the SHAP values. |
| Text classification-based approach | Patent text | Natural language processing (TF-IDF, word2vec) + ML models | Earlier stages of technology development – textual information is available | Not provided |
| The proposed approach | Patent claims | PatentHAN | | The importance of each patent claim can be obtained using a claim-wise attention mechanism |

# 3. Methodology

The overall process of the proposed approach is shown in Fig. 1. The proposed approach is designed to be implemented in four steps after patent database construction: (1) data collection and preprocessing; (2) estimation of the technological value of patents; (3) performance evaluation, and (4) prediction result interpretation.

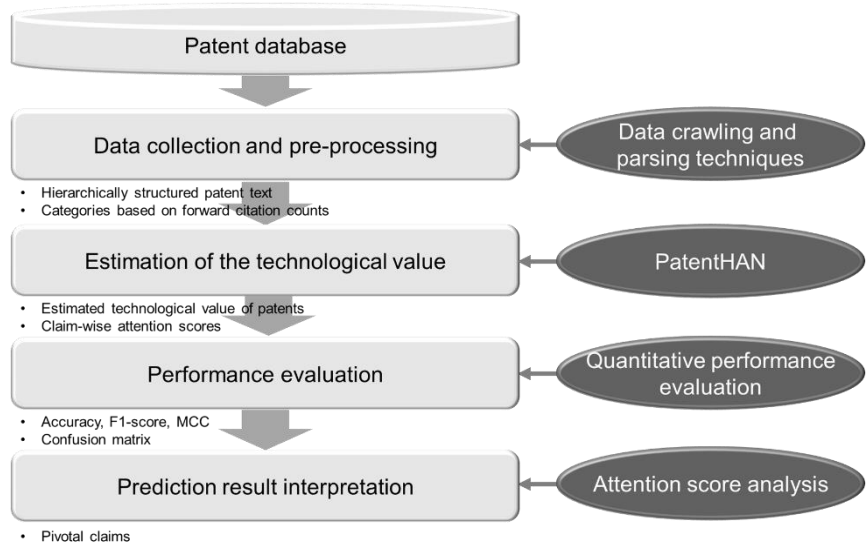

**Fig. 1.** Overall process of the proposed approach

### 3.1. Data collection and preprocessing

Once a target technology domain is determined, relevant patents can be collected from the USPTO database. The raw patent documents are usually provided in HTML or XML formats, containing a mixture of structured and unstructured data. The patents are parsed by the type of information such as patent number, class, or text, and then stored in a relational patent database for efficient management. The database should have three key categories of patent information. First, basic information comprises bibliographic information such as registration date and classification code, which can be used to define the scope and period of analysis (Choi and Yoon, 2022). Second, textual information includes patent titles, abstracts, and claims, which describe the technical contents of patents. We focus primarily on independent claims representing key technical elements of each patent. Lastly, technological value information refers to citation information (Hong et al., 2022; Lee et al., 2018), patent maintenance (Choi et al., 2020), or technology transfer (Kim et al., 2022), while their implications are different. In this study, forward citation information is utilized as a proxy for the potential of breakthrough technologies. Patent-to-patent citation relationships and citation dates are collected to estimate the number of forward citations over different observation periods.

### 3.2. Estimation of the technological value of patents.

This step aims to capture the complex meaning of technical descriptions as well as the nonlinear relationships between input and output indicators. To this end, inspired by a hierarchical attention network model (Yang et al., 2016), we develop a PatentHAN model by using PatentBERT and a transformer encoder. As shown in Fig.2., the PatentHAN model has a hierarchical structure; (1)

Beginning from the token-level encoder, words of patent claims are transformed into dense embedding vectors using PatentBERT, generating individual claim vectors. (2) In the claim-level encoder, the interactive relationships between claim vectors are captured using a claim-wise self-attention mechanism, generating a patent representation vector. (3) In the prediction layer, the patent vector is forwarded to the feed-forward neural network layer for patent class prediction. Detailed implementations of each component are as follows.

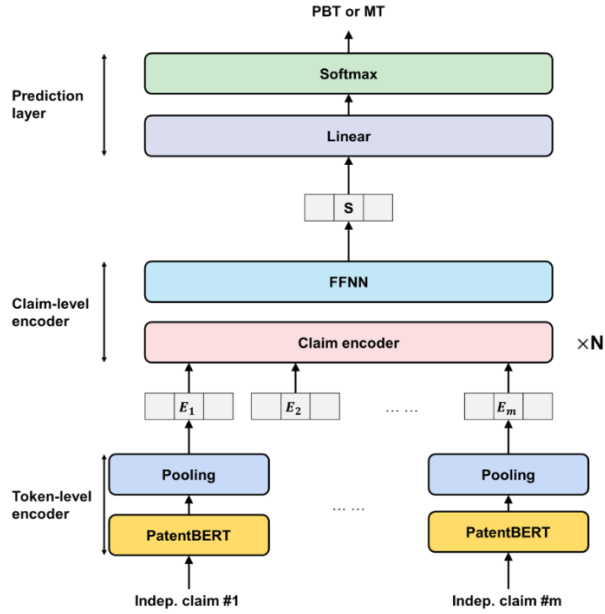

**Fig. 2.** Graphical illustration of the proposed model

This component generates dense embedding vectors, considering the complex meaning of patent texts. Specifically, token-level encoders recognize tokens from patent claims, generate their embedding vectors using PatentBERT, and provide a claim vector by averaging its constituent token vectors. Here, PatentBERT, distinguished for its advanced performance across various patent natural language processing tasks, is employed as the main PLM of our token encoder. Its exceptional comprehension of patent text, encompassing semantic and syntactic relationships between words, stems from fine-tuning the BERT model on large-scale patent datasets. Moreover, given that the token recognition ability is depending on the training corpus, PatentBERT is indispensable for precise word identification in patent texts. The resulting output from this component is detailed as follows:

$$P = (E_1, E_2, \dots, E_m) \hspace{3cm} \text{Equation (1)}$$

where $E_i$ represents the i[th] claim vector, with dimension of $d_e$, obtained by averaging token embeddings within each claim using the token-level PatentBERT encoder. Here, *m* denotes the maximum number of claims set by the users, with zero padding applied if the number of claims is less than *m*.

Claim-level encoder: The claim-level encoder extracts claim structure information to generate an intermediate patent representation, *S*, using claim vectors, *P*. As shown in Fig. 3, this step utilizes the stacked claim encoders and a feedforward neural network with tanh activation (Devlin et al., 2018; Vaswani et al., 2017). In each claim encoder, the single-head self-attention mechanism is employed and the output is calculated as Equation 2 and 3. Here, claim-wise attention mechanism enables the model to concentrate on the most semantically significant claims, by assigning different weights to each claim. Once trained, the attention mechanism is already computed during the model's training and inference processes; no additional steps are required to interpret machine predictions.

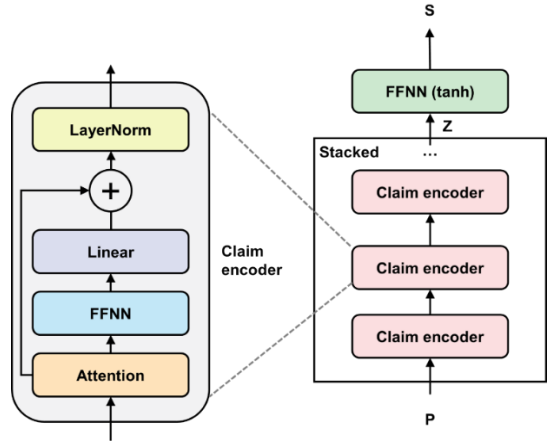

**Fig. 3.** Graphical illustration of stacked claim encoders

$$\text{Att}(P) = \text{LayerNorm}(P + \text{SingleHead}(P)) \qquad \text{Equation (2)}$$

$$\text{SingleHead}(P) = \text{head} \times W^O \qquad \text{Equation (3)}$$

where LayerNorm denotes the process of layer normalization. $W^O \in \mathbb{R}^{d_e \times d_e}$ signifies the weight matrix responsible for dimension transformation. The value of the head is calculated as follows:

$$\text{head} = Attention(Q, \text{K}, \text{V}) = Softmax(Q \times K^T / \sqrt{d_e}) \times \text{V} \qquad \text{Equation (4)}$$

$$\text{Q} = P \times W^Q, \text{K} = P \times W^K, \text{V} = P \times W^V \qquad \text{Equation (5)}$$

Here, $W^Q$, $W^K$, and $W^V$ represent the query, key, and vector weight matrices of the single head, respectively. The SoftMax function denotes an m×m matrix, where the entry at the a$^{th}$ row and

j$^{th}$ columns refers to the attention weight that the a$^{th}$ claim pays to the j$^{th}$ claim. In this context, $V$ holds the claim information for the subsequent layers and the Softmax layer serves as a gate, which impedes claims with low attention scores from propagating their information extensively. Finally, the output of the attention layer is fed into a feed-forward neural network with the residual mechanism and layer normalization as Equation 6.

$$P' = \text{LayerNorm}\,(\text{Att}(P) + Gelu(Att(P) \times W^r) \times W^S) \qquad \text{Equation (6)}$$

Here, the matrix $W^r$ serves as a transformation weight matrix, altering the dimensionality of Att(P), whereas $W^S$ is a weight matrix incorporating dropout to revert the raw dimensionality. Subsequently, $P'$ is fed into multiple claim encoders to construct the matrix Z, utilized for computing the intermediate patent representation S in Equation 7. After mean pooling, $Z$ is multiplied by $W^t$ and fed into the hyperbolic tangent function, commonly employed in neural networks for nonlinear transformations.

$$S = \text{Tanh}(\text{Avg}(Z) \times W^t) \qquad \text{Equation (7)}$$

Prediction layer: Finally, the prediction layer is added to the claim-level encoder to determine whether the given patent is a potential breakthrough technology or marginal technology. The output of the claim-level encoder ($S$) is utilized to calculate the raw score for each class, as denoted in Equation 8, where the scores for potential breakthrough technologies ($t_{PBT}$) and marginal technologies ($t_{MT}$) are fed into a SoftMax for computing loss.

$$[t_{PBT}, t_{MT}] = S \times W \qquad \text{Equation (8)}$$

## 3.3. Evaluation of model performance

The performance of the proposed approach and its ability to classify patents according to their expected technological value classes are carefully evaluated for multiple metrics. Accuracy is used to reflect the overall correctness of a model's predictions, essentially quantifying the proportion of correctly classified instances (Equation 9). Precision refers to the ratio of the correctly labeled positive examples to the total examples labeled as positive (Equation 10), whereas recall represents the ratio of the correctly labeled positive examples to the total number of positive examples (Equation 11). In this case, precision pertains to the proportion of predictions that are correctly identified as potential breakthrough technologies out of all the predictions labeled as potential breakthrough technologies.

Similarly, recall refers to the proportion of all actual breakthrough technologies that are correctly identified as such, out of the total number of potential breakthrough technologies in the dataset.

$$\text{ACCURACY} = \frac{(TP+TN)}{(TP+TN+FP+FN)}.$$   Equation (9)

$$\text{PRECISION} = \frac{TP}{TP+FP}.$$   Equation (10)

$$\text{RECALL} = \frac{TP}{TP+FN}.$$   Equation (11)

where, true positive (TP) represents examples that are correctly identified as potential breakthrough technologies, whereas false positive (FP) refers to examples that are incorrectly labeled as potential breakthrough technologies.

These metrics indicate how well a model performs in terms of correct predictions; however, it may not be the most suitable metric in cases of imbalanced datasets. For cases of imbalanced datasets, where the number of patents corresponding to potential breakthrough technologies is less than that of marginal technologies, other metrics such as MCC can be effective (Chicco and Jurman, 2020). MCC is a well-balanced metric that assesses the classifier performance by considering the TP, TN, FP, and FN (Equation 12). Its value ranges from –1 to 1, with '1' indicating perfect classification, '0' indicating no better than random classification, and '–1' indicating complete disagreement between predicted and actual values.

$$\text{MCC} = \frac{(TP \times TN) - (FP \times FN)}{\sqrt{(TP + FP)(TP + FN)(TN + FP)(TN + FN)}}$$   Equation (12)

### 3.4. Interpretation of model predictions

Once the best-performing model is developed, the predictions for a set of patents can be obtained. For instance, if a patent is predicted as a potential breakthrough technology, it is expected to have the potential for ground-breaking innovation in the future. Meanwhile, patents that are predicted as marginal technologies are likely to have less technological impact on follow-up patents. In addition to the predictions, we can identify the relative importance of each claim for the prediction in a quantitative manner, using claim-wise attention scores. As indicated in Equation 4, SoftMax serves as the attention weight matrix utilized to compute the weight of each claim for an input patent and we use the matrix of the last claim encoder for interpretation. Specifically, during training, the claim-wise attention weights are updated through the backpropagation of errors. Thus, the model learns to assign higher attention weights to claims that contribute more to the final prediction, consequently improving the model's ability to capture pivotal claims of each patent.

# 4. Empirical analysis and results

## 4.1. Overview

We conducted a case study of pharmaceutical technology for several compelling reasons. Firstly, in the pharmaceutical industry, an individual patent frequently serves as a direct representation of a product, thus establishing a strong connection between the technological merits of the patent and its commercial value (Chen and Chang, 2010). Secondly, given the relative ease with which manufacturing processes can be replicated in this field and the potential for replication with only a fraction of the original investment, the importance of rigorous patent management is pronounced compared to other sectors (Chaudhuri, 2005). Finally, there exists a demand in practice for objective and reliable information based on scientific methods to forecast the prospects of pharmaceutical technology development, due to the substantial investment and high risks of R&D activities. Hence, failure in decision-making can result in significant financial losses for companies, and unfortunately, such failures are not uncommon (DiMasi et al., 2003). In this context, the proposed method can effectively narrow down the pool of potential breakthrough technologies by filtering marginal technologies in the early stages of technology development.

## 4.2. PatentHAN approach to early screening potential breakthrough technologies

### 4.2.1. Data collection and preprocessing

A total of 35,376 patents in the field of pharmaceutical technology were obtained from the USPTO for a reference period spanning 10 years (2000–2009). Information including the patent numbers, publication years, patent claims, and count of forward citations were extracted from these patents and subsequently stored in our relational patent database. Technological value classes were defined based on the count of forward citations received within 3, 5, and 10 years after patent registration (Table 2). Following previous studies (Hong et al., 2022; Woo et al., 2019), the patents that received a large number of forward citations within the specific period, essentially exceeding the top 10%, were categorized as potential breakthrough technologies. The results of labeling is not reported here in its entirety owing to lack of space, but part of them is shown in Table 3.

**Table 2.** Definition of technological value class based on forward citation counts

| Class | Short-term forecasting | Mid-term forecasting | Long-term forecasting |
|---|---|---|---|
| Potential breakthrough technologies | Greater than or equal to 3 (10%) | Greater than or equal to 7 (8%) | Greater than or equal to 18 (12%) |
| Marginal technologies | 0–2 (90%) | 0–6 (92%) | 0–17 (88%) |

**Table 3.** Parts of the technological values of patents

| Patent Number | # of forward citations in 3 years | # of forward citations in 5 years | # of forward citations in 10 years | Category (Short-term) | Category (Mid-term) | Category (Long-term) |
|---|---|---|---|---|---|---|
| 6010682 | 0 | 1 | 6 | MT | MT | MT |
| 6010686 | 0 | 0 | 3 | MT | MT | MT |
| 6010700 | 3 | 5 | 11 | PBT | MT | MT |
| 6010709 | 7 | 7 | 10 | PBT | PBT | MT |
| 6013628 | 2 | 6 | 21 | MT | MT | PBT |
| … | | … | | … | … | … |
| 7429646 | 1 | 2 | 2 | MT | MT | MT |
| 7438910 | 0 | 8 | 20 | MT | PBT | PBT |
| 7452530 | 2 | 5 | 5 | MT | MT | MT |
| 7635568 | 6 | 12 | 22 | PBT | PBT | PBT |
| 7635487 | 4 | 7 | 14 | PBT | PBT | MT |

**Note:** PBT and MT refer to potential breakthrough technologies and marginal technologies, respectively.

For the pre-processing of independent claims, first, their stop words and claim index were removed, the text was lowered, and accent representations were removed. Next, the maximum length of tokens for each claim was set as 512. Note that the maximum number of claims should be determined in structuring each patent into a unified form for computational complexity and generalized performance. We set it as 18 according to the top 20% of the number of independent claims, thus implying that most of the patents had less than 18 independent claims. Patents with claims fewer than this value were processed with padding, whereas patents with more claims were cut to preserve the preceding claims in order.

4.2.2. Estimation of the technological value of patents

First of all, it is noteworthy that the hyperparameters, such as learning rate and batch size, should be carefully determined. This exhaustive task often relies on the size and the nature of the data set. Although models with complex structures and numerous parameters may exhibit high performance in typical tasks such as natural language processing, computer vision, or speech recognition, these tasks can lead to overfitting issues, ultimately resulting in poor generalized performance. We tested, several learning rates (1e-4, 5e-5, 3e-5, 2e-5, 1e-5, 1e-6) and batch sizes (64, 128, 256, 512), and the optimal learning rate and batch size were determined as 2e-5 and 512, respectively. The epochs were set to 100, and the early stopping technique was adopted to avoid overfitting and enable efficient learning. For the development of the PatentHAN model, we used PyTorch and HuggingFace packages and set the number of claim encoders to 4. For training details, the cross-entropy loss function and Adam optimizer were employed during the training process. Under this setting, our models were trained to classify patents according to the expected forward citation counts over the next 3, 5, and 10 years immediately after patent registration. That is, patents were divided into potential breakthrough technology or marginal technology groups based on citation counts, and then random sampling was applied within each group with a sampling rate proportional to the population of each group. Here, we used a stratified 5-fold cross-validation technique to accurately evaluate the imbalanced datasets. Also, 80% of the total patents were used as the training dataset, while the remaining 20% were used as the test dataset.

Parts of the results are given in Table 4, due to lack of space. The last three columns indicate the predicted value categories for the test dataset. Patents exhibit diverse dynamics, with many falling under the category of marginal technologies. Given that only a minority of patents receive multiple citations, while the majority remain uncited throughout their lifespan, it would be natural that most patents were predicted as marginal technologies. For instance, patent 6010682 ('Liposoluble compounds useful as magnetic resonance imaging agents') was predicted as marginal technologies for all forecasting horizons, whereas patent 7645568 ('Xenograft heart valves') was identified as breakthrough technologies for all forecasting horizons. Interestingly, patents 6013628 ('Method for treating conditions of the eye using polypeptides') and 7438910 ('Therapeutic human anti-IL1-R1 monoclonal antibody') were initially regarded as marginal technologies for short-term forecasting but as potential breakthrough technologies for long-term forecasting, which implies that they have minimal impacts at the outset of their life cycles but gain recognition as valuable technologies after undergoing market verification processes.

**Table 4.** Results of estimation of technological values of patents

| Patent Number | Number of Independent claims | Category (Short-term forecasting) | Category (Mid-term forecasting) | Category (Long-term forecasting) |
|---|---|---|---|---|
| 6010682 | 6 | MT | MT | MT |
| 6010686 | 6 | MT | MT | MT |
| 6010700 | 5 | MT | MT | MT |
| 6010709 | 9 | PBT | PBT | MT |
| 6013628 | 8 | MT | MT | PBT |
| … | … | … | … | … |
| 7429646 | 6 | MT | MT | MT |
| 7438910 | 21 | MT | MT | PBT |
| 7452530 | 11 | MT | MT | MT |
| 7635568 | 33 | PBT | PBT | PBT |
| 7635487 | 6 | PBT | MT | MT |

### 4.2.3. Evaluation of model performance

The validity of the models was investigated based on their ability to screen potential breakthrough technologies (Table 5). We employed a stratified five-fold cross-validation technique to accurately assess the imbalance datasets in our case study. The accuracies for the short-term, mid-term, and long-term periods were 0.895, 0.917, and 0.886, respectively. In addition, we computed the precision, recall, F1-score, and MCC of each class by forecasting horizon. Overall, it is observed that the proposed approach is effective in filtering patents with marginal technological values, thereby reducing the set of potential breakthrough technologies. The obtained recall values indicated that our model successfully identified a significant portion of marginal technologies across all forecast horizons. The proposed approach provided somewhat conservative predictions for potential breakthrough technologies. The performance in this category was relatively low, but the F1 scores and MCC indicated that the proposed approach is more effective compared with random classifications. The precision values for the short-term, mid-term, and long-term forecasts of breakthrough technologies were approximately 46.3%, 50.6%, and 53.8%, whereas the recall values were approximately 16.8%, 14.6%, and 22.1%, respectively.

**Table 5.** Performance evaluation of the proposed approach

**(a) short-term forecasting**

| | Potential breakthrough technologies | Marginal technologies | Overall |
|---|---|---|---|
| Accuracy | − | − | 0.895 |
| Precision | 0.463 | 0.910 | 0.687 |
| Recall | 0.168 | 0.977 | 0.573 |
| F1-score | 0.246 | 0.942 | 0.594 |
| MCC | − | − | 0.232 |

**(b) mid-term forecasting**

| | Potential breakthrough technologies | Marginal technologies | Overall |
|---|---|---|---|
| Accuracy | − | − | 0.917 |
| Precision | 0.506 | 0.927 | 0.717 |
| Recall | 0.146 | 0.987 | 0.567 |
| F1-score | 0.227 | 0.956 | 0.592 |
| MCC | − | − | 0.240 |

**(c) long-term forecasting**

| | Potential breakthrough technologies | Marginal technologies | Overall |
|---|---|---|---|
| Accuracy | − | − | 0.886 |
| Precision | 0.538 | 0.904 | 0.721 |
| Recall | 0.221 | 0.975 | 0.598 |
| F1-score | 0.313 | 0.938 | 0.623 |
| MCC | − | − | 0.294 |

4.2.4. Interpretation of model predictions

In addition to the predictions regarding potential breakthrough technologies or marginal technologies, a claim-wise attention score was calculated for each prediction (Fig. 4.). Examining independent claims with comparatively high attention scores elucidates the reasons behind the predictions, whether the given patent is breakthrough technologies or not. For instance, as shown on the left side of Fig. 4., the patent (registration number: 6664289) was predicted as a potential breakthrough technology, its 24th claim is recognized as a pivotal claim with the highest score of 0.556. The invention pertains to an aqueous nasal solution designed for treating and preventing microbial infections and encompasses a broad-spectrum microbicide, including aqueous chlorine or bromine, hypochlorite ion, and/or chloride, bromide, or iodide ion. In this context, the 24th claim was recognized as the most pivotal, as it provides a wide spectrum of antimicrobial agents for the treatment and prevention of various microbes, which are not specified in other claims. In addition, this claim encompasses various concentrations of chemical substances, implying that the invention can be adjusted to suit specific environments or purposes. Also, by specifying concentration ranges for each chemical substance, this claim helps meet safety regulations and compliance with regulatory agencies, which plays a crucial role in verifying the safety and efficacy of the relevant product.

**Fig 4.** Example of claim-wise attention scores.

On the right side of Fig. 4, for the patent (registration number: 7588786) predicted as marginal technologies, its independent claims and corresponding attention scores are provided. The given invention pertains to a eutectic-based self-nano emulsified drug delivery system, which is primarily utilized for administering poorly water-soluble drugs to patients. This claim describes ubiquinone dissolution under 37 degrees Celsius using ubiquinone and volatile essential oils, but its broad scope may diminish patentability and value. In addition, lacking practical information for the in vivo delivery system could reduce patent utility. In summary, the flexible and extendable purpose and scope such as materials and conditions in the independent claims is observed in the patents in potential breakthrough technologies. Therefore, the provision of clear evidence or key technical elements rather than merely providing lengthy and verbose content is likely to determine the technological scope of patents and further their potential values.

# 5. Discussion

5.1. Validation of the proposed approach

The outcomes and implications of the proposed approach may vary according to the analysis context. In this regard, we performed further experiments to examine the effects of different encoding layers or claim types, which are adaptable to the given context. First, we examined the impact of using other PLMs as encoding layers in the proposed approach. Specifically, we employed PLMs such as BERT (Devlin et al., 2018), SciBERT (Beltagy et al., 2019), BioBERT (Lee et al., 2020), and PubMedBERT (Gu et al., 2021), of which training datasets are different. Table 6 shows the robust performance of the proposed approach using other PLMs. It was observed that SciBERT is the most effective for the three forecasting tasks, exhibiting slightly improved performance compared to the baseline model. This is because the target domain is the pharmaceutical field, where understanding the given textual information requires scientific knowledge, even though the data type is patent. Therefore, when using the proposed approach in practice, it is important to test various PLMs trained on data types or domains similar to the target domain.

**Table 6.** Impact of different PLMs as a text encoding layer of PatentHAN

| PLMs | Short-term forecasting | | Mid-term forecasting | | Long-term forecasting | |
|---|---|---|---|---|---|---|
| | Accuracy | MCC | Accuracy | MCC | Accuracy | MCC |
| PatentBERT (Baseline) | 0.895 | 0.232 | 0.917 | 0.240 | 0.886 | 0.294 |
| BERT | 0.895 | 0.169 | 0.909 | 0.165 | 0.876 | 0.228 |
| SciBERT | 0.894 | 0.247 | 0.917 | 0.250 | 0.889 | 0.317 |
| BioBERT | 0.892 | 0.204 | 0.915 | 0.238 | 0.886 | 0.294 |
| PubMedBERT | 0.894 | 0.226 | 0.916 | 0.240 | 0.876 | 0.302 |

Secondly, we examined the effects of incorporating dependent claims as additional input data. Patents consist of two types of claims: independent claims and dependent claims. An independent claim outlines the fundamental technical aspects of an invention and establishes its scope, while a dependent claim elaborates on specific technical details, relying on an independent claim for context. Table 7 provides the performance of the proposed method when only independent claims are used compared to when both independent and dependent claims are utilized. It was observed that utilizing only independent claims as input yields higher performance in terms of accuracy and MCC, compared to incorporating both types of claims. This is because independent claims convey more pivotal information, while dependent claims provide specific and verbose details. Therefore, we can conclude that the existence of pivotal technical elements or functions may be crucial in determining the potential of breakthrough technologies.

**Table 7.** Impact of different claim types as input data of PatentHAN

| PLMs | Short-term forecasting | | Mid-term forecasting | | Long-term forecasting | |
|---|---|---|---|---|---|---|
| | **Accuracy** | **MCC** | **Accuracy** | **MCC** | **Accuracy** | **MCC** |
| Independent claims only | 0.895 | 0.232 | 0.917 | 0.240 | 0.886 | 0.294 |
| Dependent claims added | 0.862 | 0.088 | 0.915 | 0.115 | 0.869 | 0.127 |

In addition, we tried to examine the statistical difference between the importance of independent and dependent claims by comparing the attention scores by claim type. For this, we used the models that use both independent claims and dependent claims and analyzed the attention scores which were normalized by patent. Specifically, we conducted a two-tailed t-test for unequal sample size and unequal variance. The null hypothesis was $Y_1 = Y_2$ whereas the alternative hypothesis was $Y_1 \neq Y_2$, where $Y_1$ and $Y_2$ indicate the mean value of attention scores for the independent claims and dependent claims. As shown in Table 8, the findings revealed significant differences between the two claim sets, further corresponding to prior works arguing that independent claims hold paramount importance (Marco et al., 2019).

**Table 8.** Summary of *t*-test results of claim-wise attention scores depending on the claim type.

| | Short-term forecasting | Mid-term forecasting | Long-term forecasting |
|---|---|---|---|
| *t*-statistic | 31.98 | 31.73 | 31.29 |
| Degree of freedom | 53,017 | 53,757 | 52,788 |
| Mean of scores in group 1 | 0.8213 | 0.8504 | 0.8264 |
| Variance of scores in | 0.0731 | 0.0562 | 0.0705 |

| | group 1 | | |
| --- | --- | --- | --- |
| Mean of scores in group 2 | 0.7108 | 0.7531 | 0.7192 |
| Variances of scores in group 2 | 0.0708 | 0.0577 | 0.0668 |
| *p*-value | 0.0 | 0.0 | 0.0 |

**Note:** Group 1 and group 2 indicate the groups of independent claims and dependent claims.

## 5.2. Implications for academia and practice

The case study showed that the proposed approach is effective for early screening potential breakthrough technologies from patent text, providing enhanced interpretability. Therefore, the proposed approach is expected to have significant implications for both theory and practice. From an academic perspective, this study contributes to technology innovation research by suggesting an interpretable ML method for early screening of potential breakthrough technologies. This is made through the development of PatentHAN, enabling the identification of pivotal claims in the screening process. Unlike previous methods based on opaque ML models, the proposed approach incorporates the claim-wise attention scores, thereby enhancing experts' understanding of the model predictions. Although the focus of this study is an early screening of potential breakthrough technologies, the proposed approach can be applied to solve other research questions where the comprehension and interpretation of technical contents are important. These may include but are not limited to technology valuation (Kim et al., 2023), technology opportunity analysis (Choi et al., 2023; Seol et al., 2023), inventor scouting (Chung et al., 2021) and patent-trademark linking (Ko et al., 2020). From a practical perspective, the proposed approach is expected to advance the timing of screening potential breakthrough technologies, as it requires only textual descriptions of technologies as input. Therefore, given sufficient technical descriptions, proof-of-concept or prototype-stage technologies can be evaluated with the proposed approach. The proposed approach and interpretable results can significantly aid practitioners in making informed decisions regarding breakthrough technologies while facilitating effective communications between stakeholders.

## 6. Conclusion

This study has proposed an interpretable ML approach for early screening of potential breakthrough technologies. Central to this approach is the development of a PatentHAN model, which provides interpretable results in predicting the number of forward citations from patent text. The core components of this model include: (1) PatentBERT is utilized as text encoding layer to covert patent claim texts into dense embedding vectors that capture their intricate semantic meanings; (2) a hierarchical network structure, where individual claim vectors are organized to create a comprehensive patent representation vector, is designed to facilitate detailed claim-level analysis; and (3) a claim-wise attention mechanism, which enables the model to concentrate on the most semantically significant parts by assigning weights to each claim, is employed to effectively interpret machine predictions by identifying pivotal claims during the screening process. The case study of pharmaceutical technology confirmed that the proposed approach is effective in screening potential breakthrough technologies and claim-wise attention scores support machine-experts collaborations by revealing the most pivotal technical elements from patent texts.

Despite its effectiveness, this study is subject to several limitations. First, the proposed approach utilizes the number of forward citations as a single proxy of technological value. However, incorporating other indicators such as technology transfer and patent renewal data is required in future works, for comprehensive evaluation of potential breakthrough technologies (Choi et al., 2020; Kim et al., 2022; Ko et al., 2019). Second, the screening performance can be further improved with the latest models. Although we used an appropriate PLM for patent text processing, the latest large language models such as PaLM2 (Anil et al., 2023), and GPT-4 (Achiam et al., 2023) could provide more accurate predictions. Third, quantitative indicators on the market, technology or regulation factors or image data of patents can be used as additional input data (Choi et al., 2022; Choi et al., 2021; Jee et al., 2022), and the development of multi-modal models could potentially enhance the accuracy and reliability (Chung and Sohn, 2020). Lastly, this study solely focused on a single case study on pharmaceutical technology. To establish the external validity of our approach, further testing should be performed on technologies across diverse domains. Nonetheless, we argue that the proposed approach and interpretable results make significant contributions to both academia and practice.